\RequirePackage{fix-cm}
\documentclass[smallextended]{svjour3}       
\smartqed  

\usepackage{graphicx}
\usepackage{graphics}
\usepackage{subfigure}
\usepackage{graphicx}
\usepackage{amsmath}
\usepackage{cite}
\usepackage{booktabs}
\usepackage[table]{xcolor}
\usepackage{tabularx}
\usepackage{url}

\def \1{\columnwidth}
\def \2{0.48\columnwidth}
\def \3{0.3\columnwidth}
\def \4{0.25\columnwidth}
\def \5{0.2\columnwidth}
\def \75{0.8\columnwidth}
\begin{document}

\title{Enhanced image feature coverage: 
}
\subtitle{Key-point selection using genetic algorithms}

\titlerunning{Enhanced image feature coverage}        

\author{Erkan Bostanci}


\institute{F. Author \at
              Computer Engineering Department, Ankara University, Golbasi Campus, 06830, Ankara,
              Turkey \\
              Tel.: +90-312-203-3300 pbx. 1767\\              
              \email{ebostanci@ankara.edu.tr}           
          }


\maketitle

\begin{abstract}
Coverage of image features play an important role in many vision algorithms since their distribution affect the estimated homography. This paper presents a Genetic Algorithm (GA) in order to select the optimal set of features yielding maximum coverage of the image which is measured by a robust method based on spatial statistics. It is shown with statistical tests on two datasets that the metric yields better coverage and this is also confirmed by an accuracy test on the computed homography for the original set and the newly selected set of features. Results have demonstrated that the new set has similar performance in terms of the accuracy of the computed homography with the original one with an extra benefit of using fewer number of features ultimately reducing the time required for descriptor calculation and matching.

\end{abstract}

\section{Introduction}
\label{sec:introduction}

Efficiency and accuracy of the feature detection and matching process are very important considerations in various vision applications such as object recognition and tracking as well as 3D reconstruction. Recent work~\cite{Bostanci2014a} has shown that the distribution of the image features\footnote{Image features and key-points are used in the text interchangeably.} across the image plays an important role in this process since it is known that this distribution affects homography estimation.

If the locations of detected image features are analysed, one can see that possible clusterings of features are likely to occur in the vicinity of specific image features depending on the scene content like texture, corners, edges, \emph{etc.} This situation can get more severe if  multi-scale detectors (\emph{e.g.} SIFT~\cite{Lowe2004}) are employed since these detectors find the same feature at different levels of the scale space. With such inhomogeneities present, first-order metrics such as density of the features are not capable of describing the variation between these feature clusters and result in an incomplete or overlooked description of the image coverage.

When the literature is examined, it can be seen that this problem, indeed, attracted a significant amount of attention by different researchers aiming to measure, use or improve coverage. Some intuitively believed that well-distributed feature matches should yield to more accurate homography estimations \cite{Hartley2003, Sola2007, Ma2004}. Sola~\cite{Sola2007} explicitly addressed this problem in his thesis by placing a notional grid over the image and trying to select similar numbers of Harris corner features from each tile in order to obtain a uniform feature distribution. An approach that is very similar to this was given in~\cite{Ma2004} where Harris features were computed separately from each tile of a similar grid, a computationally more expensive method. 

Yet another approach is very well known and inherent to the feature detection process itself, namely Non-Maximum Suppression (NMS)~\cite{Nixon2002,Neubeck2006}. This approach is a step in computer vision algorithms. Recent examples involve the extraction of interest points as in SUSAN~\cite{Smith1997}, FAST~\cite{Rosten2010} and SFOP~\cite{Foerstner2009}. NMS basicly defines a point saliency measure computed for the whole image and/or in the scale space and then extracts the local maxima, aiming to prevent feature clusterings at particular regions. 

A number of approaches to measure feature coverage across an image can also be seen in the literature. Perdoch et al.~\cite{Perdoch2007} presented an explicit definition of coverage through a qualitative evaluation which uses a frame detection method to analyse what portion of the image was covered by detected image features. The work in~\cite{Dickscheid2009} employed the convex-hull as the measure of spatial coverage and Tuytelaars et al.~\cite{Tuytelaars2010} proposed dense sampling method for improving coverage which used inter-feature spatial relationships. The work by Ehsan et al.~\cite{Ehsan2011} employed the harmonic mean of feature distances as the coverage metric since this approach penalized some of the feature detectors that yield feature clusters. It is also possible to find studies~\cite{Dickscheid2011, Aanaes2011, Ehsan2013} which analyse complementarity of feature detectors (\emph{i.e.} employing different detectors together) to achieve a better feature coverage. Recently, a metric based on spatial analysis techniques, Ripley's K-function~\cite{Ripley1976, Ripley1977} in particular, was developed in~\cite{Bostanci2014a, Bostanci2012a}. This developed metric both presents a quantitative measure as opposed to visual inspection approach~\cite{Perdoch2007} and takes the variations in the feature density across the image into account rather than averaging or convex hull approaches in~\cite{Crawley2007, Bivand2008, Dickscheid2011}.

The focus of this paper is on finding the optimal set of features that achieve better image coverage due to its practical importance in vision applications. The work in~\cite{Bostanci2014a} found that SFOP (similar results were found by~\cite{Dickscheid2009}) was giving good coverage which also has good invariance and repeatability characteristics, comparable to that of SIFT~\cite{Lowe2004}. The work presented will propose a GA that will maximize the feature coverage by eliminating clusterings of features from the output of SFOP. A work uses Ant Colony Optimization to reduce image features can be found by Chen et al.~\cite{Chen2011}, which focuses on features for image classification which are first/second order moments, entropy \emph{etc.}, though image features mentioned here are key-points computed by feature detectors. Another work is given in~\cite{Ozkan2006} which aims to select the best SIFT key-points for face recognition, \emph{i.e.} ones giving best matches; however, no experimental results were given. Here, the set of features giving optimal coverage are selected and statistical analysis on the accuracy of the estimated homography is performed in order to check whether the new set of features can yield the similar results with the original set of features, \emph{i.e.} to see if statistically significant differences would be found even when using a smaller number of features. 

The rest of the paper is outlined as follows: Section~\ref{sec:coverageMetric} describes the coverage metric presented as an improvement which is followed by Section~\ref{sec:featureSelection} where the feature selection algorithm based on genetic algorithms is proposed. The evaluation of the algorithm is given in Section~\ref{sec:evaluation} and then the results are presented in Section~\ref{sec:results}. Finally, conclusions are drawn in Section~\ref{sec:conclusion}.

\section{Coverage Metric}
\label{sec:coverageMetric}

In spatial analysis, point distributions are classified into three patterns, namely: regular (dispersed), random and aggregated~\cite{Crawley2007}. The first distribution presents a uniform distribution of key-points over the image. Random aggregations are likely to occur in a random pattern, generally following a Poisson distribution. The last distribution exhibits even more clusters.

Ripley's K-function ($K(r)$) provides a good description of the presence of clustering of feature points in an image as demonstrated by~\cite{Bostanci2014a}. $K(r)$ is a function of distance, it is able to describe the density of feature points at many distance ($r$) scales.  If there are $N$ key-points within area $A$ and the distance between any two key-points $i$ and $j$ is $r_{ij}$, then one can estimate $K(r)$~\cite{Ripley1976} as
\begin{equation}
\hat{K}(r) = \frac{A}{N^2}\sum_j\sum_{I, i\not=j} \frac{I_r(r_{ij})}{w_{ij}}
\label{eq:kEstimate}
\end{equation}
where $I_r(\cdot)$ and $w_{ij}$ determine whether a point will be
included in the calculation at radius $r$ and whether points lying on
the boundaries of $A$ are counted~\cite{Crawley2007}.

\begin{figure}[h!t!b!p]
	\begin{center}    
		\subfigure[Dispersed and clustered points]
		{\includegraphics[width=\2]{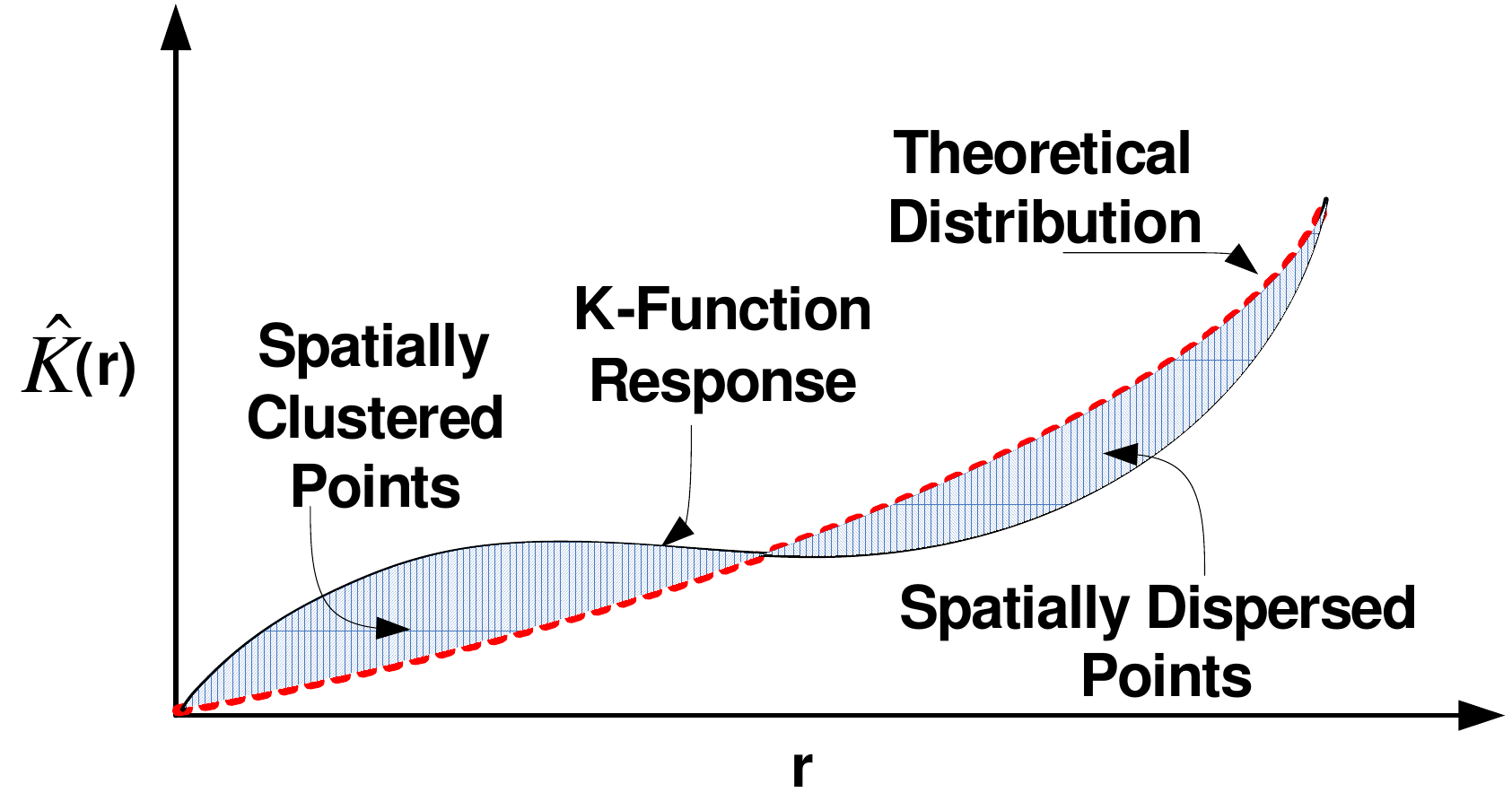}}
		\subfigure[Definitions used in the new metric]
		{\includegraphics[width=\2]{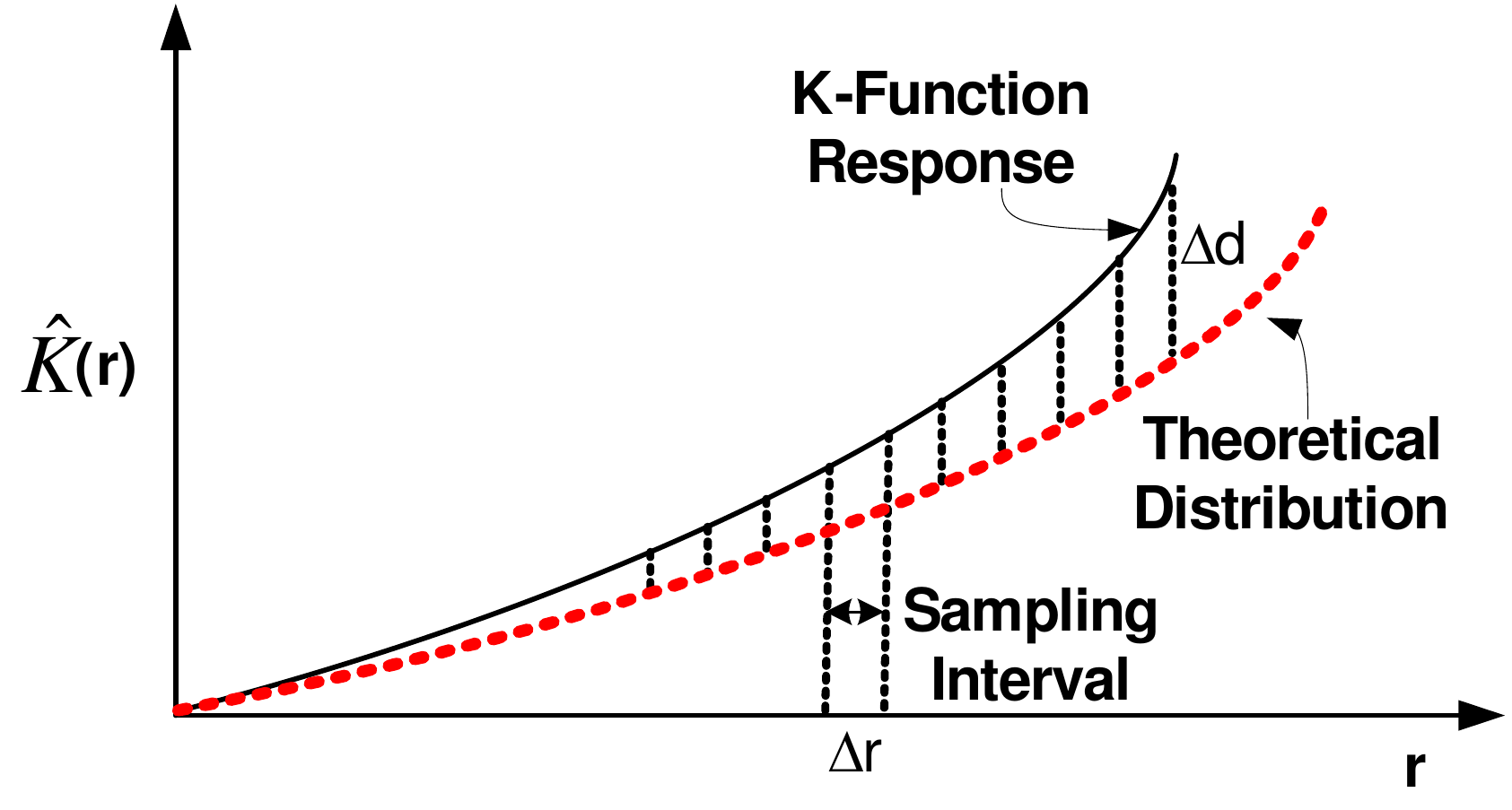}}    
	\end{center}
	\caption{Ripley's K function and the proposed coverage metric}
	\label{fig:kExplanations}
\end{figure}

For a homogeneous Poisson process, one expects~\cite{Dixon2002}
\begin{equation}
K_P(r) = \lambda \pi r^2
\label{eq:kTheoretical}
\end{equation}
Fig.~\ref{fig:kExplanations}-a depicts a plot of $K_P(r)$  against experimentally-measured $\hat{K}(r)$ calculated with (\ref{eq:kEstimate}). Regions in the figure where $\hat{K}(r) > K_P(r)$ show that the image features are aggregated for the given distance of $r$, while values where $\hat{K}(r) < K_P(r)$ indicate regular distributions of image features. It is important to emphasize that the main aim here is not to fit Poisson distribution to the features but to identify whether the features are dispersed or clustered since the former is desirable for an accurate estimate of the homography.

The work in~\cite{Bostanci2014a} employed the difference of Area Under the Curve (AUC) for the $\hat{K}(r)$ and $K_P(r)$ as the coverage metric. Here, this metric is selected as the difference between selected values of $r$ for a specific sampling interval denoted with $\Delta r$ (Fig.~\ref{fig:kExplanations}-b). The main reason for this is the computational efficiency, since the former approach (difference of AUCs) required more processing. With this consideration in mind, the new coverage metric is defined as:
\begin{equation}
\alpha = \sum\limits_{r_{min},\Delta r}^{r_{max}} \left|K(r) - K_P(r)\right|
\label{eq:newMetric}
\end{equation}

It is important to note here that when working with image features, one should not always expect to have results below $K_P(r)$ due to the content of the image, \emph{e.g.} non-uniformities in the content or texture. Hence, clusterings are more likely to occur rather than a regular distribution. For this reason, it would be adequate to minimize $\alpha$ making it as close possible to the theoretical distribution to achieve better coverage.

The value of $\alpha$ stands as the basis of the feature selection approach using a genetic algorithm in order to achieve a better feature coverage. 

\section{Feature Selection Using GA}
\label{sec:featureSelection}

\begin{figure}[h!t]
	\begin{center}    
		\includegraphics[width=\75]{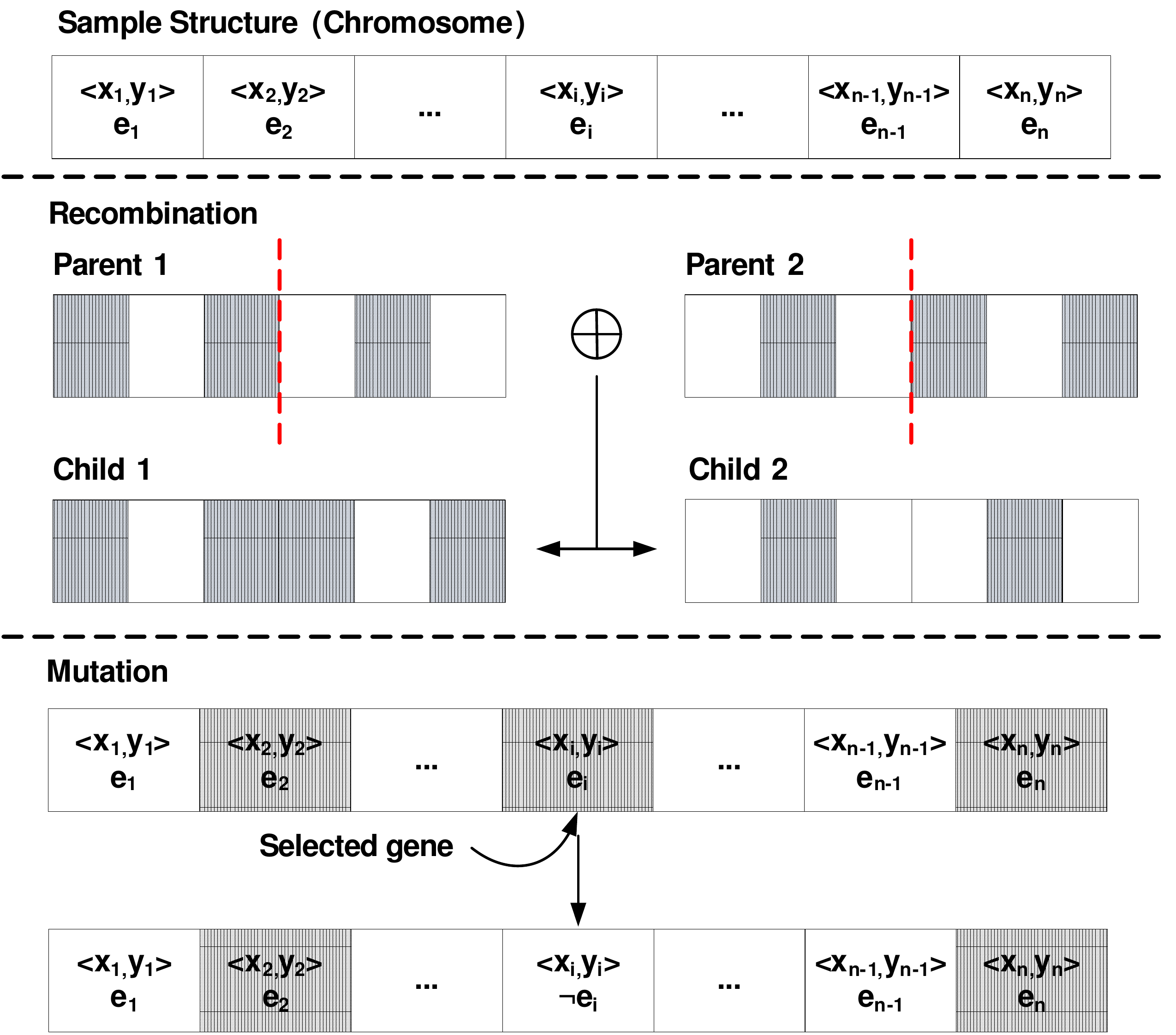}   
	\end{center}
	\caption{Sample structure and genetic operators}
	\label{fig:geneticOps}
\end{figure}

The aim for maximizing image feature coverage mentioned in the introduction is an example of optimization problems for which GAs have been employed frequently~\cite{Eiben2003}. The problem can be described as extracting an optimal set of image features $F'$ from the complete set of features $F$ that will yield the maximum coverage which can be denoted as:
\begin{equation}
\min_{\left<f_1, f_2, \ldots, f_n \right> \in F}~\alpha(f_1, f_2, \ldots, f_n) 
\label{eq:optimization}
\end{equation}
A GA makes use of exploitation and exploration operators, recombination and mutation respectively, in order to look for the optimal result in the search space. The optimal result here is the set of features producing optimal coverage. A population of individuals (samples) are employed in GA and genetic operations on these individuals are performed in order to find the fittest one. A candidate set of image features is represented by a sample with the structure shown in Fig.~\ref{fig:geneticOps}.

Each image feature $f_i$ is represented with its image coordinates $\left(x_i,y_i\right)$ as well as a single boolean variable $e_i$ which indicates whether the feature will be included in the computation of the coverage metric defined in Section~\ref{sec:coverageMetric}. This metric was employed as the fitness of a sample for the evaluation stage. The evolutionary process will eventually yield the optimal set of features with maximum coverage ($\alpha$ will be minimized). 

Selection was performed using the roulette-wheel algorithm~\cite{Goldberg1989} in which a random number is generated and then compared against the accumulated (normalized) fitness of the individuals, the first individual with a fitness value higher than the generated value is selected for recombination for choosing both parents to perform 10 cross-overs producing 20 siblings added to the population. The maximum size of the population was chosen as 100 with an initial population size of 10.

Recombination was implemented as one-point cross-over as depicted in Fig.~\ref{fig:geneticOps}. Experiments with two-point cross-over did not produce satisfactory results which may be due to the fact that feature detector uses a scanning algorithm to add key-points  to the final result, \emph{i.e.} starting from one corner of the image to the opposite one. 

Mutations were used to explore the undiscovered areas of the search space and implemented as simple negation on $e_i$ ($\neg e_i$) on the allele of the gene which was selected for a mutation with a rate of $0.030$ (see Fig.~\ref{fig:geneticOps}). Elitism was also employed in order to keep the individuals which have resulted in the best fitness values so far. 

\begin{figure}[h!t!b!p]
	\begin{center}    
		\includegraphics[width=\75]{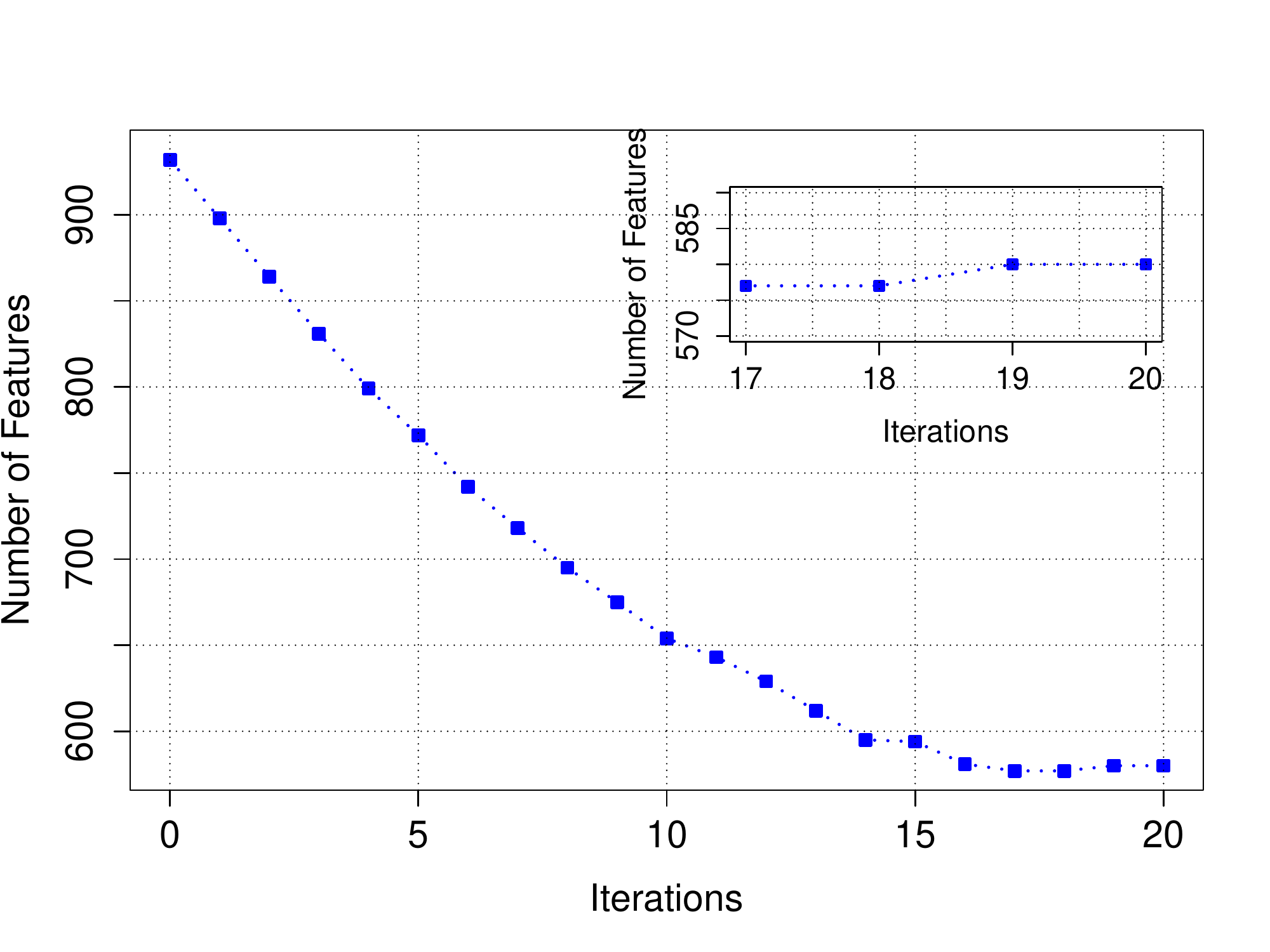} 
	\end{center}
	\caption{Change in the number of features through the evolutionary process}
	\label{fig:changeFeatureNum}
\end{figure}

Number of generations (iterations) was selected as 20 which was found to be sufficient to obtain an optimal set of features. Figure~\ref{fig:changeFeatureNum} shows the change in the number of selected features during the evolutionary process where iteration numbered 0 denotes the original set of features. The general trend is the decrease in the number of features in order to reduce the effect of clusterings on the coverage, however this is not necessarily the case. An important property of the sample design presented here is that it allows using various sets of features for distinct individuals and if required employing the feature that have been omitted in order to increase coverage. The imprint of the figure shows an example to this. Despite the overall decreasing trend in feature number, it remained stable for iterations 17 and 18; however, there is an increase in the number of features at iteration 19 and did not change in the last iteration. This dynamic structure of the algorithm not only focuses on reducing the number of features, but also aiming to increase coverage even it means increasing the number of features employed for computation.

Using the sample structure and the genetic operators presented here, the new set of features were obtained. Next step is evaluating this new set for its fitness for use in vision algorithms.

\section{Evaluation}
\label{sec:evaluation}

In order to assess the practical value of the developed approach which finds the optimal set of the image features yielding better coverage based on the presented coverage metric, an evaluation was performed using two different datasets with statistical tests. 

\subsection{Datasets}
\label{sec:datasets}

The developed metric was first tested on a newly gathered VASE dataset\footnote{Available at \url{http://vase.essex.ac.uk/datasets/index.html}} in order to assess the performance of improved feature coverage. This dataset comprises of 520 images encompassing a wide range of scene types, \emph{e.g.} indoor and outdoor scenes with a variety of illumination, texture and contrast. Original images with size $1429 \times 949$ were re-sampled to half-size using \emph{mogrify} command available in \emph{ImageMagick}~\cite{Still2005}.

A second evaluation was performed on the widely used Oxford dataset\footnote{Available at \url{http://www.robots.ox.ac.uk/~vgg/data/data-aff.html}}. This dataset includes 8 sets of paired images with various affine transformations as well as lighting changes and JPEG compression. The main reason for using this dataset is to assess the computed homography from the new set of image features and compare this with the one computed with original set of features.

\subsection{Experimental Framework}
\label{sec:experimentalFramework}

The evaluations were performed using an experimental design constructed on a \emph{null-hypothesis} framework~\cite{Bostanci2013,Bostanci2014b}. The evaluations were performed with two different criteria here. The first criterion is whether the metric is yielding better coverage for the datasets and the second one assesses whether the accuracy of the homography estimated with the new set of features is similar to the original. The performance of the coverage metric against these two criteria was the basis for this assessment. The null-hypothesis, $H_0$, here was that the performance of the new set would be similar to the original image feature set, whereas the alternative hypothesis ($H_1$) suggests that there would be statistically significant differences. 

Given a pair of images of the same scene taken from different viewpoints, there will be an overlapping region providing same features and hence feature matches. As mentioned earlier, the distribution of these features is an important parameter for the computed homography from the overlapping region. In other words, when there are clusterings of features in some parts of this region and the distribution is not mostly uniform, the computed homography will be inaccurate.

The evaluation computed homography matrices with the refined set of features and compared them with the ones computed using the original feature set. The accuracy of the homography matrices was quantified using the following approach~\cite{Bostanci2014b} (illustrated in Figure~\ref{fig:explanation}):

\begin{figure}[h!t!p]
	\centering
	\includegraphics[width=\75]{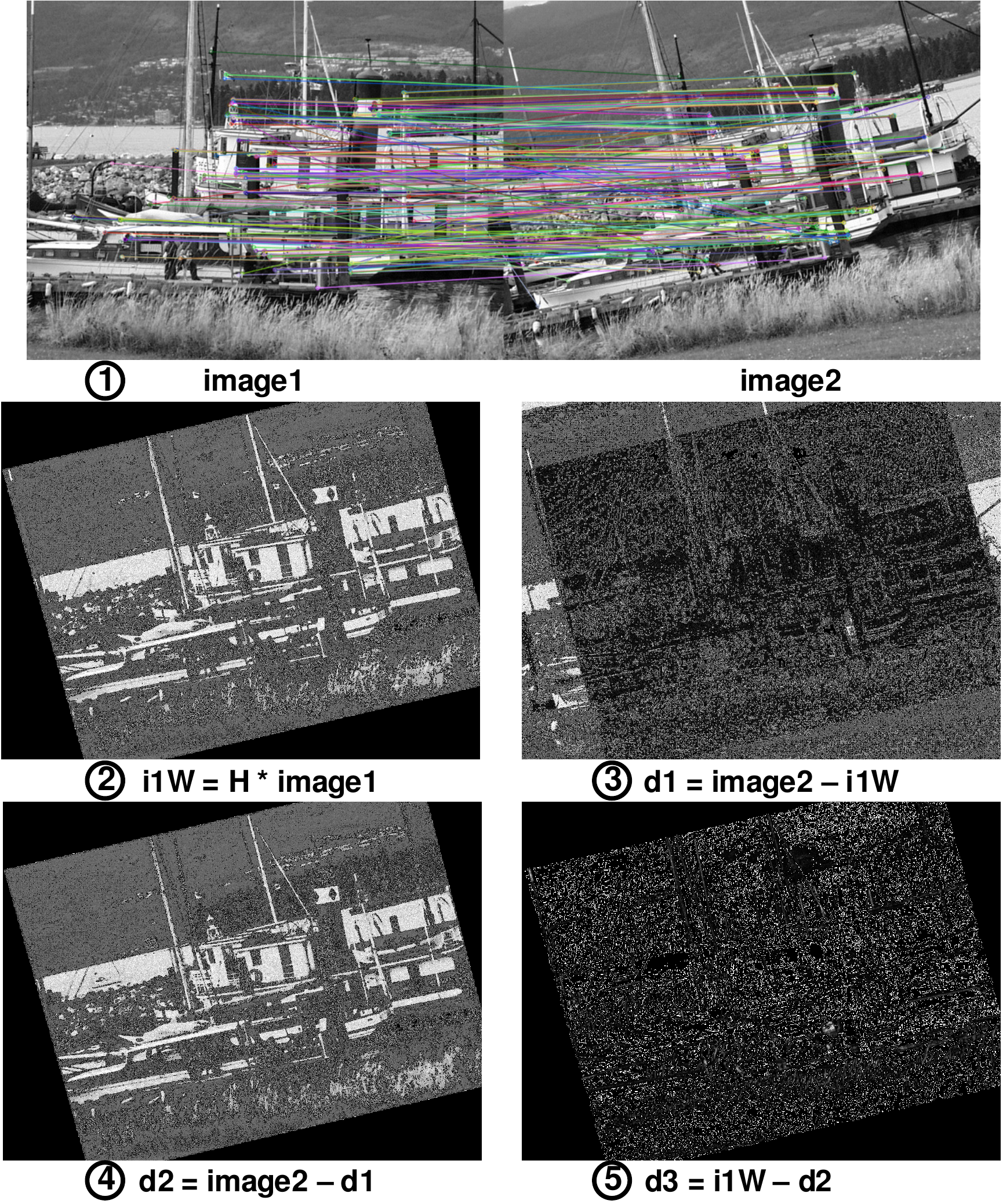}
	\caption{Process used in the evaluation}
	\label{fig:explanation}
\end{figure}

As it was mentioned earlier, image features (key-points) detected with SFOP were obtained. These features were described using the SIFT descriptor~\cite{Lowe2004} and then are matched across a pair of images ($image1$ and $image2$  in the figure). The homography matrix computed from the matching set of features was then applied to $image1$ in order to warp it onto the second, this warped image ($i1W$) is then subtracted from $image2$. The result is an intermediate image, $d1$, which is again subtracted from $image2$ to remove the non-overlapping part between $i1W$ and $image2$ yielding $d2$. Finally, a difference image $d3$ is obtained by subtracting $d2$ from $i1W$ and the number of non-zero pixels were counted in $d3$ as a measure of accuracy of the computed homography which is used as the evaluation criterion. 

Here, a larger number points to accuracy problems arising from incorrect alignment due the computed homography from the given set of image features. Both t-test~\cite{Student1908} and McNemar's test~\cite{Mcnemar1947, Clark1999} were employed in the evaluation.

\section{Results}
\label{sec:results}

This section is dedicated to the results on the evaluation described in the previous section. Table~\ref{tab:ttestCoverage} presents the t-test results on the VASE dataset and it is clear that the evolved set results in a lower value for $\alpha$ suggesting better coverage ($t$ is larger than $t$-Critical for one and two-tail tests with a $P$ value very close to zero) with substantial confidence. $H_0$ can be safely rejected. 

\begin{table}[htbp]
	\centering
	\caption{t-test results on coverage improvement using VASE dataset}
	\begin{tabular}{rrr}
		\toprule
		& \textbf{Original Set} & \textbf{Refined Set} \\
		\midrule
		\textbf{$\mu$} & 3897.1002 & 2783.8826 \\
		\textbf{$\sigma$} & 1880.8933 & 1492.5160 \\
		\textbf{Observations} & 520   & 520 \\
		\midrule
		\textbf{$t$ Stat} & 10.5723 &  \\
		\textbf{$P(T\le t)$ one-tail} & 4.0139E-25 &  \\
		\textbf{$t$ Critical one-tail} & 1.6464 &  \\
		\textbf{$P(T\le t)$ two-tail} & 8.0278E-25 &  \\
		\textbf{$t$ Critical two-tail} & 1.9624 &  \\
		\bottomrule
	\end{tabular}%
	\label{tab:ttestCoverage}%
\end{table}%

\begin{figure*}[h!t]
	\begin{center}    
		\subfigure[Original feature set]
		{\includegraphics[width=\2]{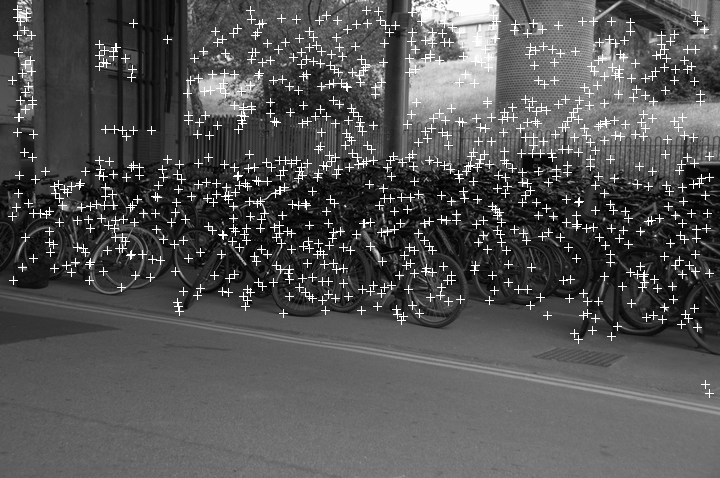}}
		\subfigure[Refined feature set]
		{\includegraphics[width=\2]{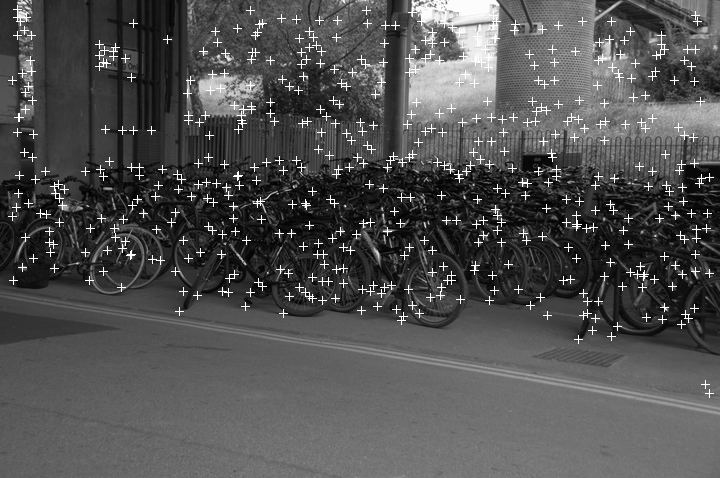}}    
		\subfigure[K-function for the original feature set]
		{\includegraphics[width=\2]{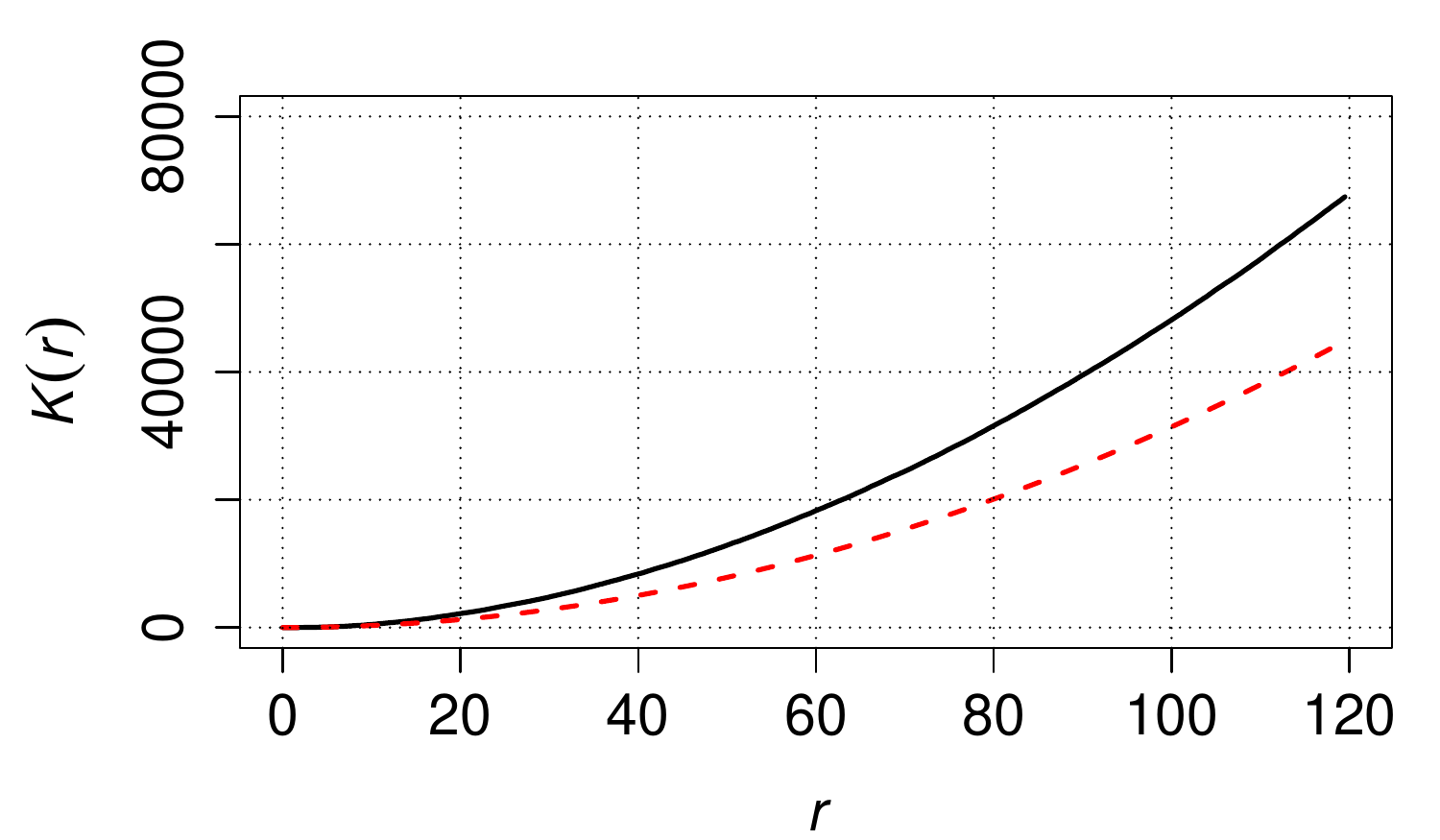}}    
		\subfigure[K-function for the refined feature set]
		{\includegraphics[width=\2]{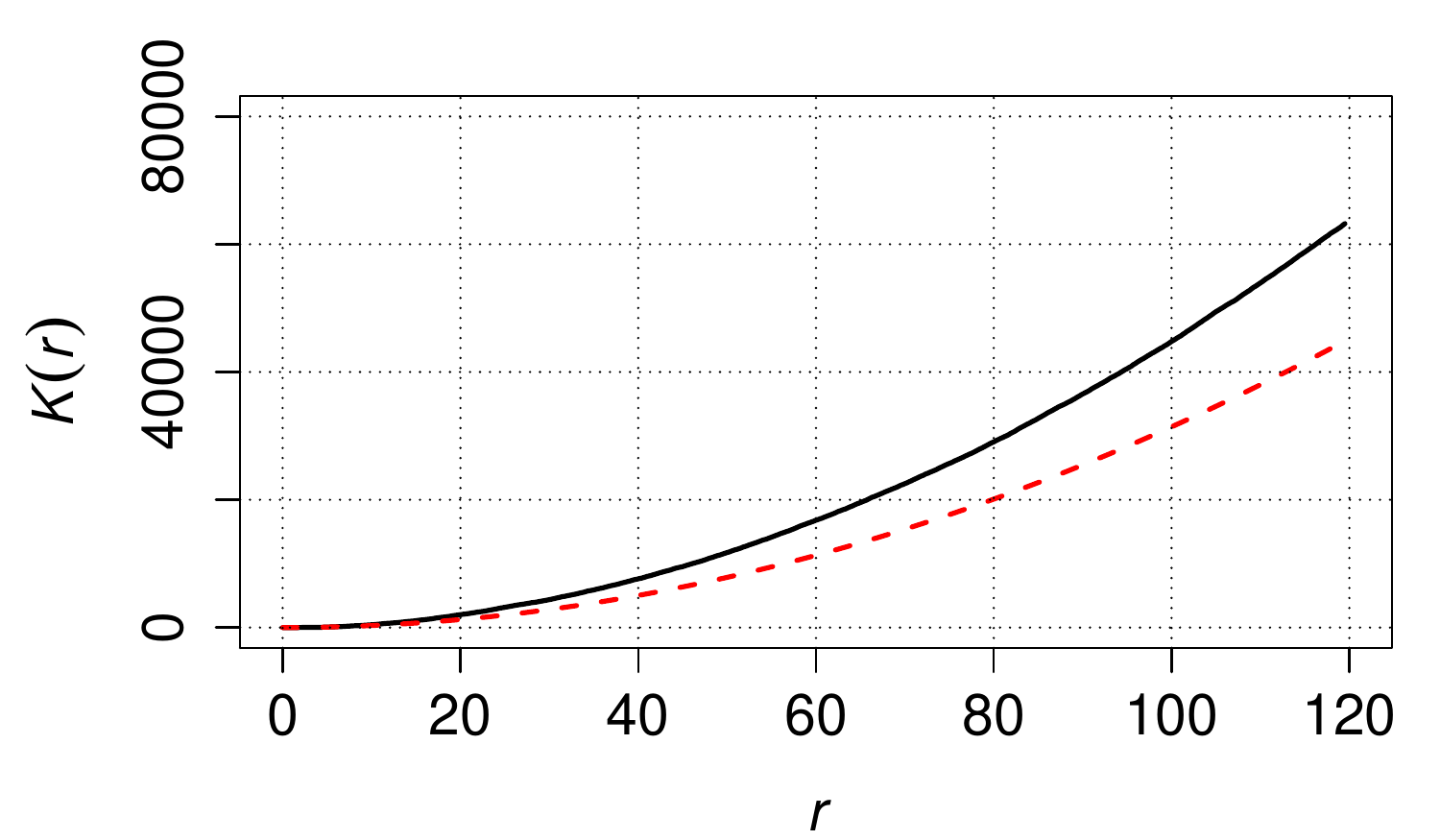}}    
		\subfigure[Grid counts for the original feature set]
		{\includegraphics[width=\2]{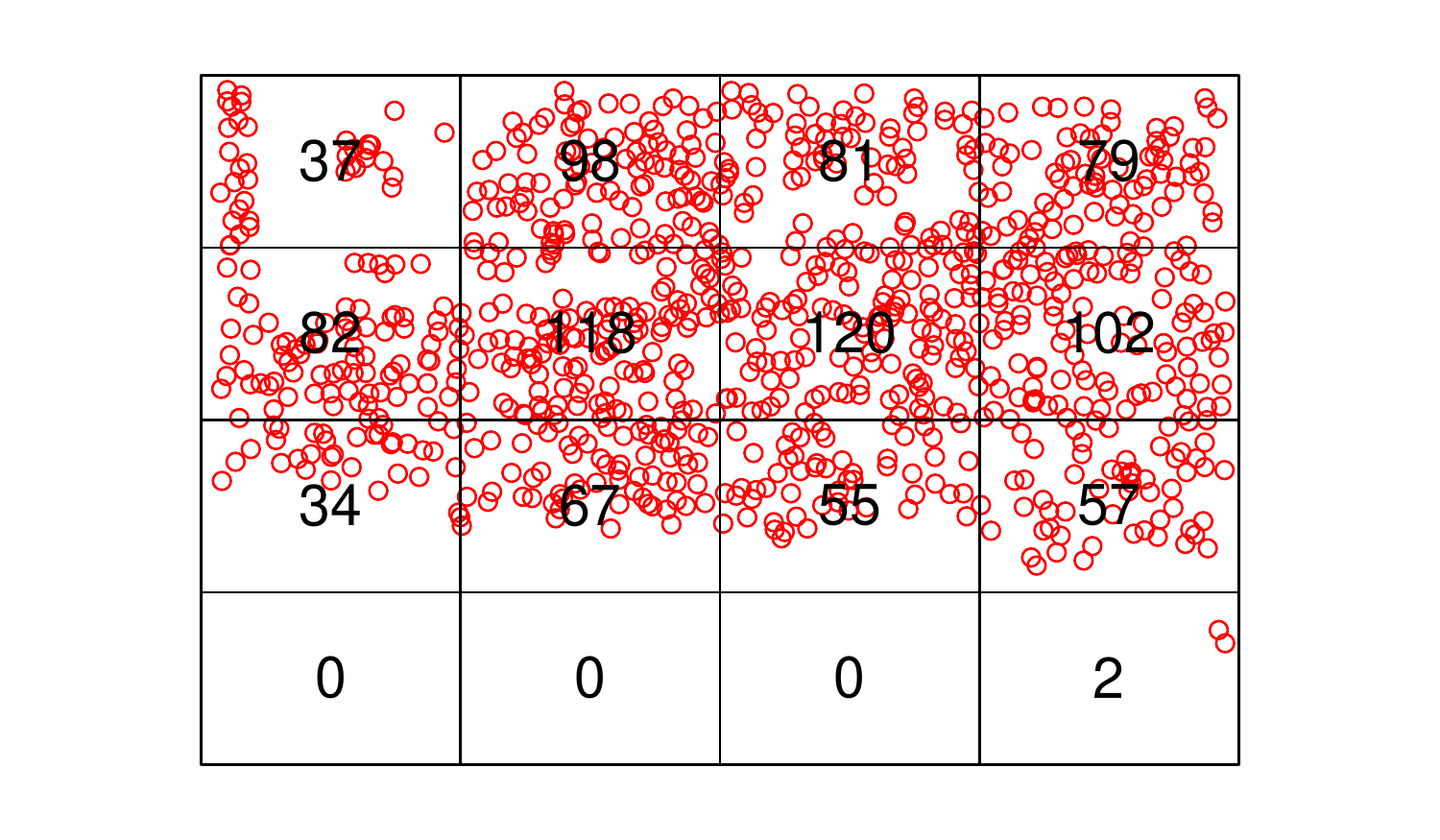}}    
		\subfigure[Grid counts for the refined feature set]
		{\includegraphics[width=\2]{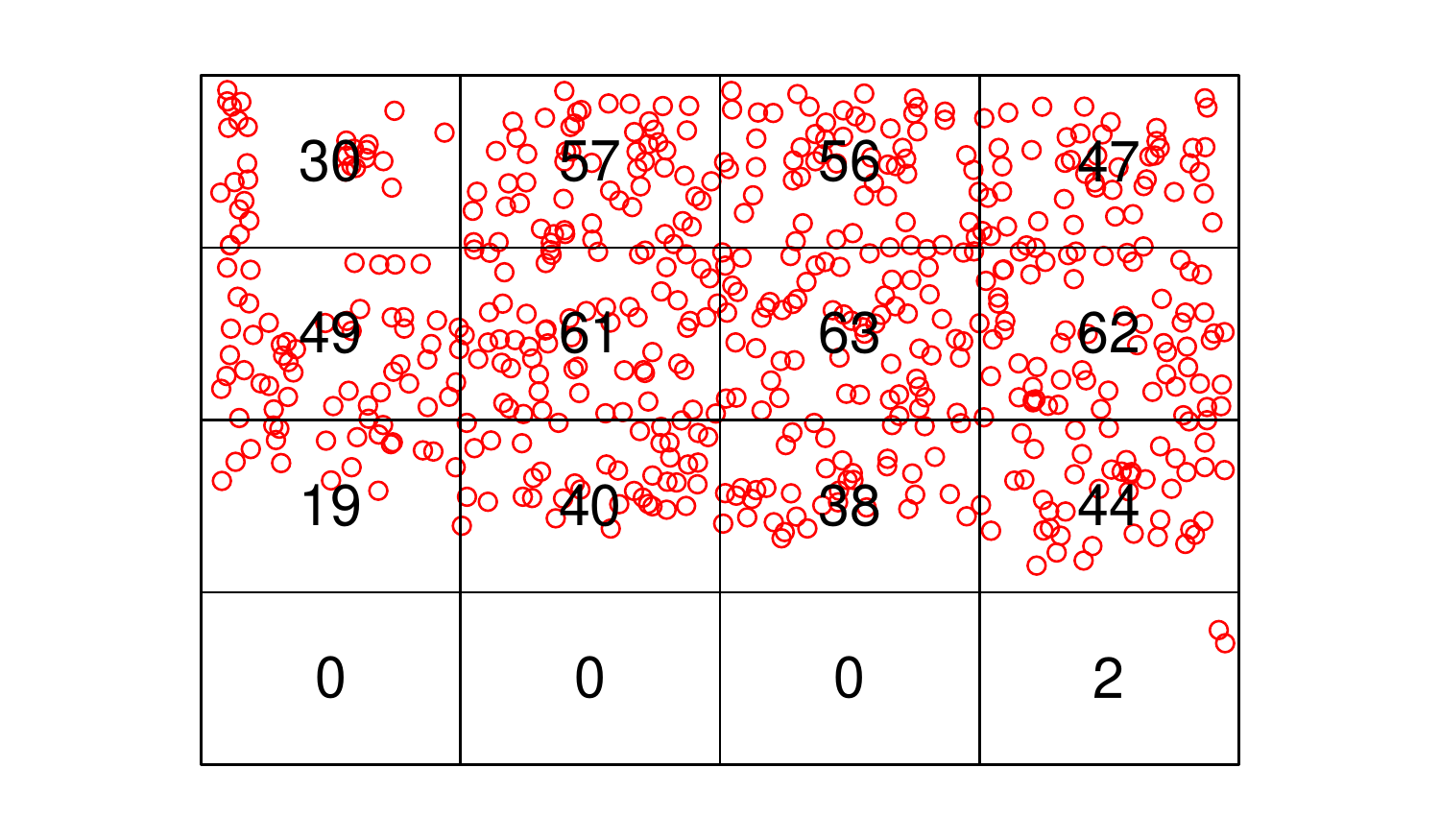}}    
	\end{center}
	\caption{Distribution of original and refined set of image features}
	\label{fig:imagesKplotsGrids}
\end{figure*}

Fig.~\ref{fig:imagesKplotsGrids} gives a view on the outcome of the algorithm. The detected set of features are shown in Fig.~\ref{fig:imagesKplotsGrids}-a and the refined set in Fig.~\ref{fig:imagesKplotsGrids}-b. A first look on these images clearly show the removal of the clusterings of features. 

How this reflects on the Ripley's K function is depicted in Fig.~\ref{fig:imagesKplotsGrids}-c and d. The difference between the experimental and the theoretical plots has decreased. A plot closer to the theoretical distribution (see Section~\ref{sec:coverageMetric}) suggests less clustering and hence better coverage. 

The grids of Fig.~\ref{fig:imagesKplotsGrids}-e and f also show the numbers of features in $4 \times 4$ tiles. What can be noticed from these is that the numbers of features are closer to each other in the new set of features (f), pointing to a more uniform distribution. Note that regions neglected by the feature detector remains the same, devoid of any key-points.

Looking at the accuracy for the estimated homography in Table~\ref{tab:ttestHomography}, it can be seen that the differences are not statistically significant $t$ is smaller than $t$-Critical with a very low confidence -- $H_0$ is not rejected suggesting similar performance in terms of the estimated homography.

\begin{table}[htbp]
	\centering
	\caption{t-test results on accuracy of computed homography}
	\begin{tabular}{rrr}
		\toprule
		& \textbf{Original Set} &  \textbf{Refined Set} \\
		\midrule
		\textbf{$\mu$}  & 4.4531 & 4.4933 \\
		\textbf{$\sigma$} & 1.6769 & 1.6807 \\
		\textbf{Observations} & 288   & 288 \\
		\midrule
		\textbf{$t$ Stat} & 0.2875 &  \\
		\textbf{$P(T\le t)$ one-tail}  & 0.3869 &  \\
		\textbf{$t$ Critical one-tail} & 1.6475 &  \\
		\textbf{$P(T\le t)$ two-tail} & 0.7739 &  \\
		\textbf{$t$ Critical two-tail} & 1.9641 &  \\
		\bottomrule
	\end{tabular}%
	\label{tab:ttestHomography}%
\end{table}%

A McNemar test also yielded a $z$ score of 0.8427 for the accuracy test, confirming the result of the t-test and suggesting no statistically significant difference in terms of performance.

\begin{figure}[h!t!b!p]
	\begin{center}    
		\subfigure[Vase dataset]
		{\includegraphics[width=\2]{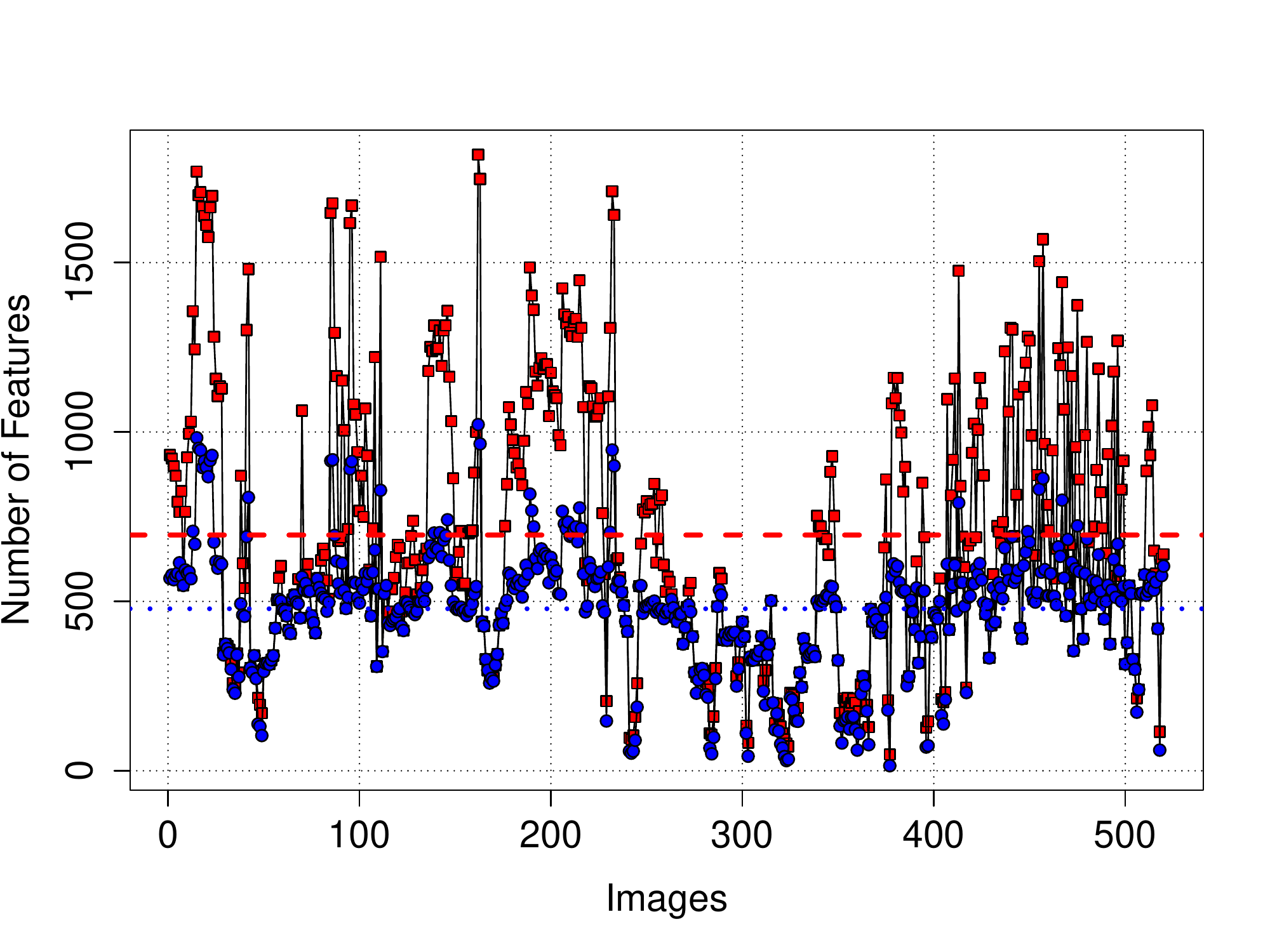}}
		\subfigure[Oxford dataset]
		{\includegraphics[width=\2]{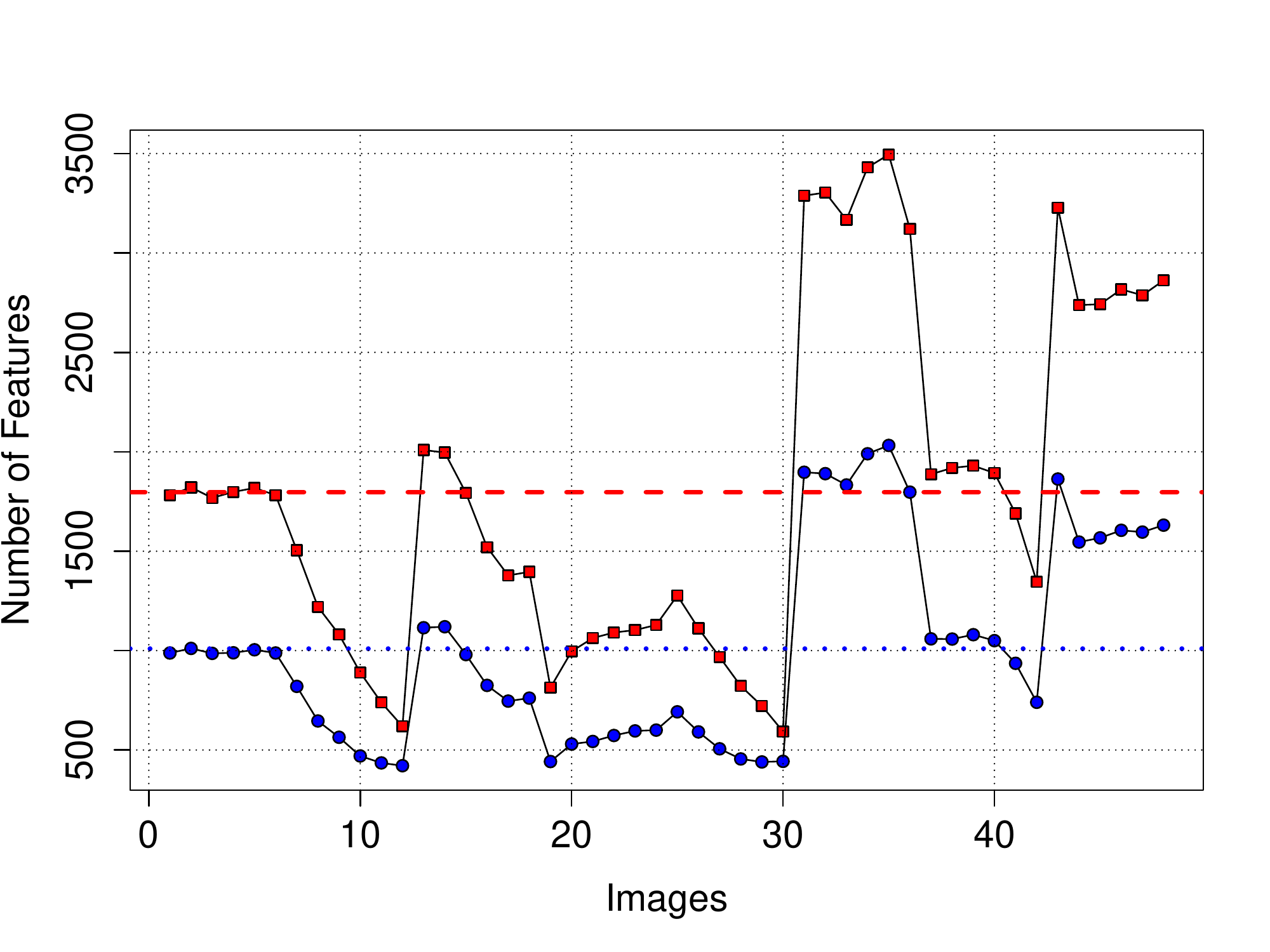}}		
	\end{center}
	\caption{Change in the number of features for original and refined set of image features, squares and circles respectively. Dashed and dotted lines represent means.}
	\label{fig:features}
\end{figure}

\begin{figure}[h!t!b!p]
	\begin{center}    
		\subfigure[Vase dataset]
		{\includegraphics[width=\2]{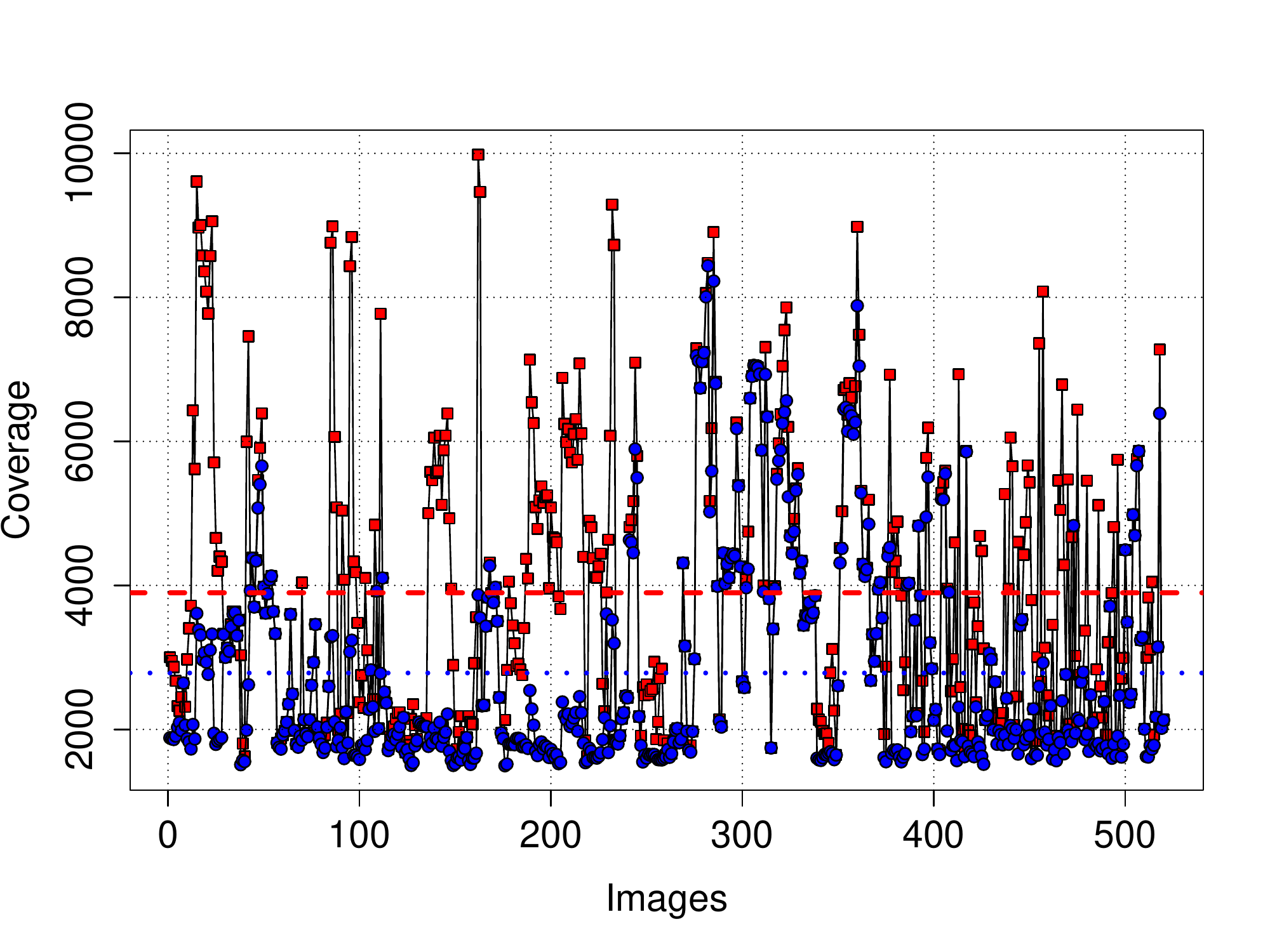}}
		\subfigure[Oxford dataset]
		{\includegraphics[width=\2]{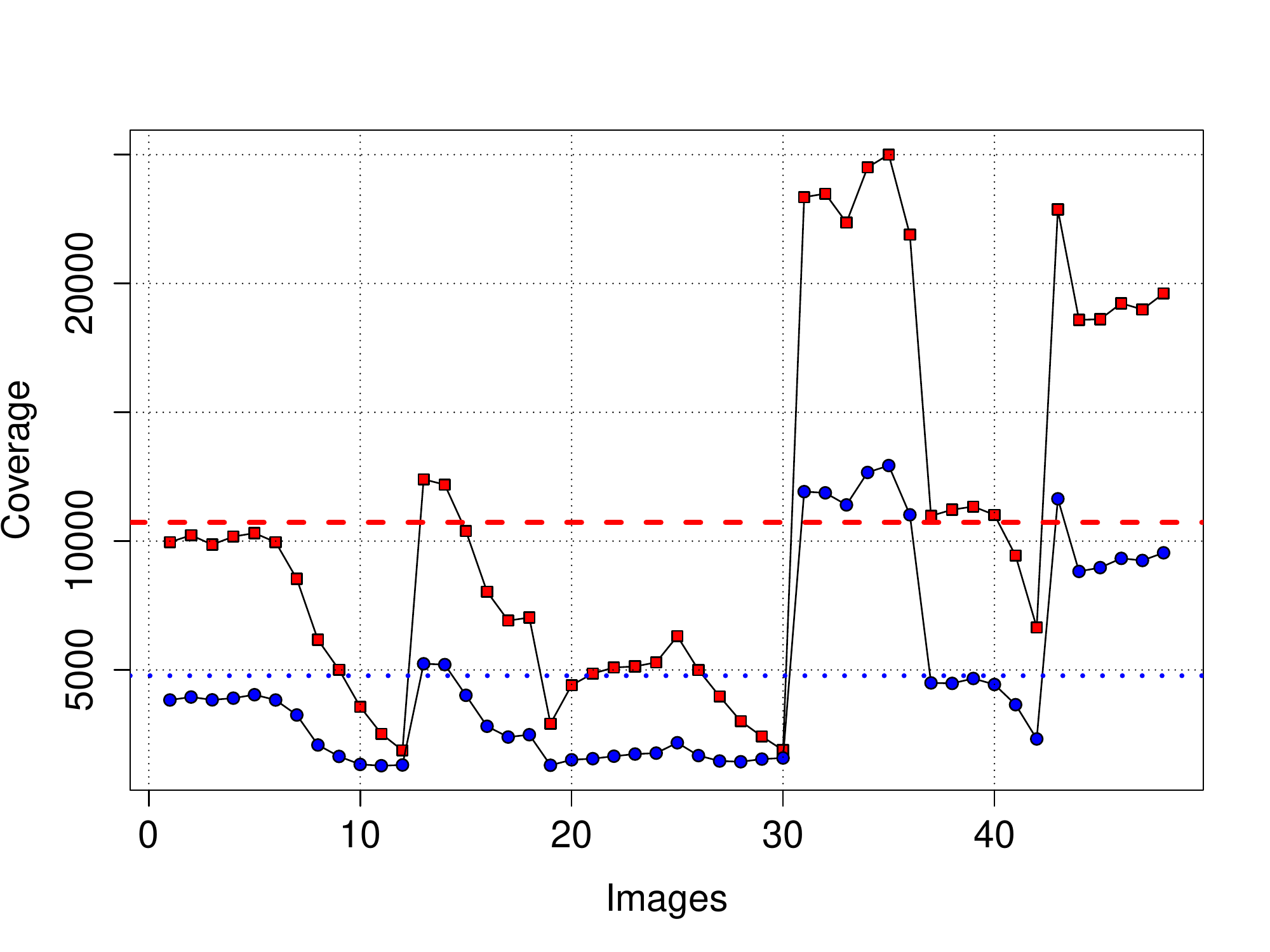}}		
	\end{center}
	\caption{Change in the coverage metric for original and refined set of image features, squares and circles respectively. Dashed and dotted lines represent means.}
	\label{fig:coverage}
\end{figure}

\begin{figure}[h!t!b!p]
	\begin{center}    
		\includegraphics[width=\75]{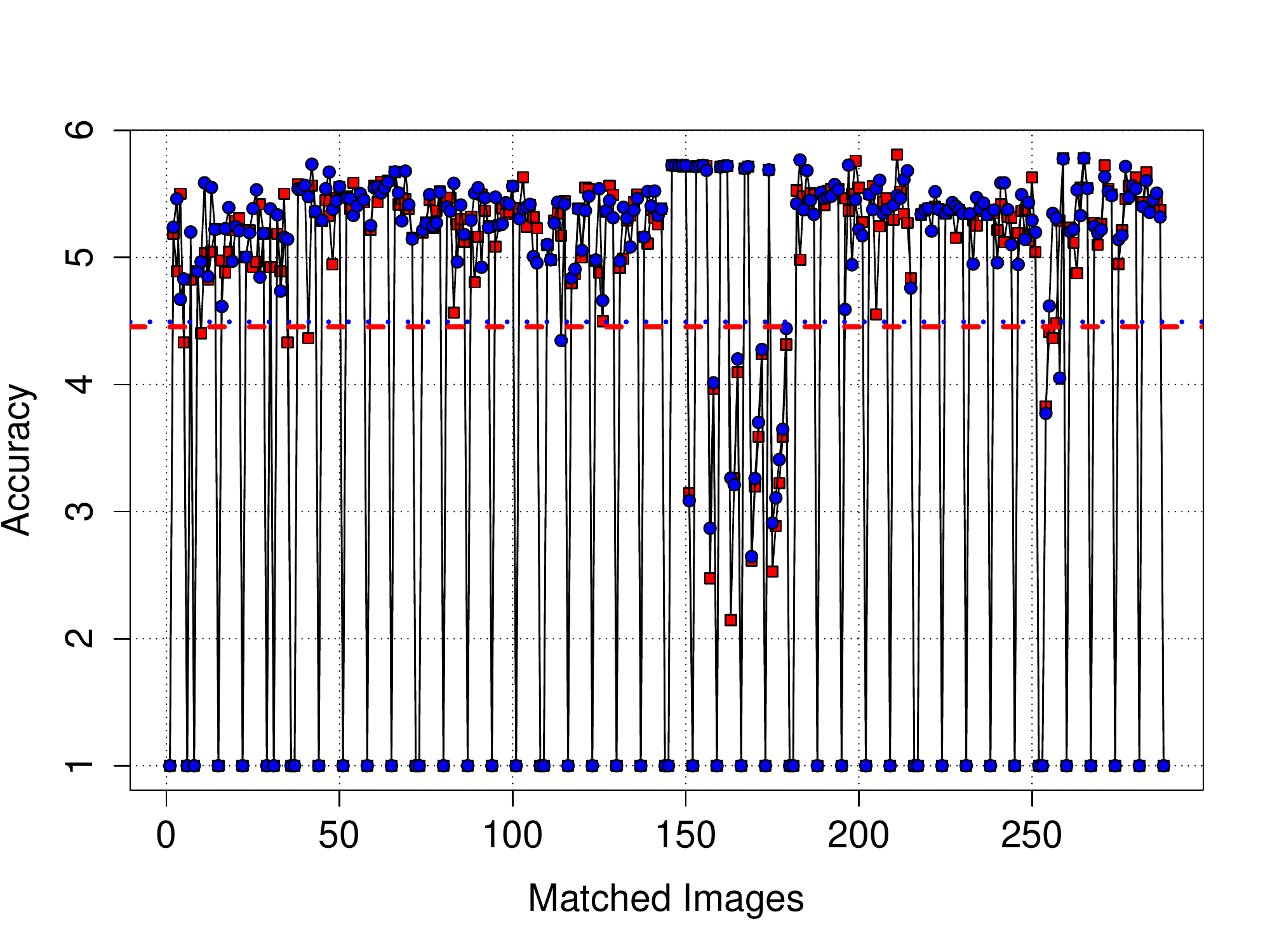}   
	\end{center}
	\caption{Change in the accuracy of the computed homography for original and refined set of image features, squares and circles respectively. Dashed and dotted lines represent means.}
	\label{fig:accuracy}
\end{figure}

Figures~\ref{fig:features} to~\ref{fig:accuracy} show the change in the number of features, image feature coverage and the differences in the computed homography respectively for the original and refined set of features. It can be noticed that no improvement was achieved for a few number of frames; however, the evolutionary algorithm has reduced the number of features (Fig.~\ref{fig:features}), improved the coverage metric (Fig.~\ref{fig:coverage}) and at the same time resulted in a close performance in terms of accuracy (Fig.~\ref{fig:accuracy}) for the majority of the dataset.

\section{Conclusion}
\label{sec:conclusion}

Many different studies have demonstrated that the distribution of image feature has an important role in the accuracy of the estimated homography. With this fact in mind, an algorithm that will receive a set of features from a feature detector's output and refine this set based on a robust coverage metric by means of a GA was presented in this paper. 

It was shown that the algorithm generally reduces the number of features by eliminating clusterings, while it was also observed that merely reducing number of features does not always improve the coverage. The algorithm chooses which features will be employed to achieve better coverage using an established metric. Though, the decrease in feature number comes as bonus, reducing the time required for descriptor calculation and matching between images.

To the best knowledge of the author's, this approach is the first one reported for refining the detector output to achieve better coverage using a recently developed statistical metric. It is completely different from NMS in which the detectors response must be known at a local scale to refine output by eliminating the non-maximum values. The proposed method does not require to know the underlying algorithm and directly works on the detector's output making it independent of the detector.

Timing concerns related to the evolutionary process inherent to GAs are vanishing gradually these days due to Graphics Processing Units (GPUs)~\cite{Pospichal2010}. Making use of this, the algorithm presented can be efficiently implemented, invalidating such concerns.

\bibliographystyle{spmpsci}      

\bibliography{References}

\begin{thebibliography}{10}
\providecommand{\url}[1]{{#1}}
\providecommand{\urlprefix}{URL }
\expandafter\ifx\csname urlstyle\endcsname\relax
  \providecommand{\doi}[1]{DOI~\discretionary{}{}{}#1}\else
  \providecommand{\doi}{DOI~\discretionary{}{}{}\begingroup
  \urlstyle{rm}\Url}\fi

\bibitem{Aanaes2011}
Aan{\ae}s, H., Dahl, A.L., Pedersen, K.S.: Interesting interest points---{A}
  comparative study of interest point performance on a unique data set.
\newblock International Journal of Computer Vision pp. 1--18 (2011)

\bibitem{Bivand2008}
Bivand, R.S., Pebesma, E., Gomez-Rubio, V.: Applied Spatial Analysis with {R}.
\newblock Springer (2008)

\bibitem{Bostanci2013}
Bostanci, B., Bostanci, E.: An evaluation of classification algorithms using mc
  nemar's test.
\newblock In: Proceedings of Seventh International Conference on Bio-Inspired
  Computing: Theories and Applications (BIC-TA 2012), \emph{Advances in
  Intelligent Systems and Computing}, vol. 201, pp. 15--26. Springer (2013)

\bibitem{Bostanci2014b}
Bostanci, E.: Is hamming distance only way for matching binary image feature
  descriptors?
\newblock Electronics Letters \textbf{50}(11), 806--808 (2014)

\bibitem{Bostanci2014a}
Bostanci, E., Kanwal, N., Clark, A.: Spatial statistics of image features for
  performance comparison.
\newblock Image Processing, IEEE Transactions on \textbf{23}(1), 153--162
  (2014)

\bibitem{Bostanci2012a}
Bostanci, E., Kanwal, N., Clark, A.F.: Feature coverage for better homography
  estimation: {A}n application to image stitching.
\newblock In: Proceedings of the IEEE International Conference on Systems,
  Signals And Image Processing (2012)

\bibitem{Chen2011}
Chen, L., Chen, B., Chen, Y.: Image feature selection based on ant colony
  optimization.
\newblock In: AI 2011: Advances in Artificial Intelligence, \emph{Lecture Notes
  in Computer Science}, vol. 7106, pp. 580--589. Springer Berlin Heidelberg
  (2011)

\bibitem{Clark1999}
Clark, A., Clark, C.: Performance evaluation in computer vision: A tutorial
  (1999)

\bibitem{Crawley2007}
Crawley, M.J.: The R Book.
\newblock Wiley-Blackwell (2007)

\bibitem{Dickscheid2009}
Dickscheid, T., F\"orstner, W.: Evaluating the suitability of feature detectors
  for automatic image orientation systems.
\newblock In: ICVS 2009, LNCS:5815, pp. 305--314 (2009)

\bibitem{Dickscheid2011}
Dickscheid, T., Schindler, F., F\"orstner, W.: Coding images with local
  features.
\newblock International Journal of Computer Vision \textbf{94}, 154--174 (2011)

\bibitem{Dixon2002}
Dixon, P.M.: Ripley's {K} function.
\newblock Encyclopedia of Environmetrics \textbf{3}, 1796--1803 (2002)

\bibitem{Ehsan2013}
Ehsan, S., Clark, A.F., McDonald-Maier, K.D.: Rapid online analysis of local
  feature detectors and their complementarity.
\newblock Sensors \textbf{13}(8), 10,876--10,907 (2013).
\newblock \doi{10.3390/s130810876}

\bibitem{Ehsan2011}
Ehsan, S., Kanwal, N., Clark, A.F., McDonald-Maier, K.: Measuring the coverage
  of interest point detectors.
\newblock In: Image Analysis and Recognition (ICIAR'11), LNCS:6753, pp.
  253--261 (2011)

\bibitem{Eiben2003}
Eiben, A.E., Smith, J.E.: Introduction to Evolutionary Computing.
\newblock SpringerVerlag (2003)

\bibitem{Foerstner2009}
F\"orstner, W., Dickscheid, T., Schindler, F.: Detecting interpretable and
  accurate scale-invariant keypoints.
\newblock In: 12th IEEE International Conference on Computer Vision (ICCV'09).
  Kyoto, Japan (2009)

\bibitem{Goldberg1989}
Goldberg, D.E.: Genetic Algorithms in Search, Optimization and Machine
  Learning, 1st edn.
\newblock Addison-Wesley Longman Publishing Co., Inc., Boston, MA, USA (1989)

\bibitem{Hartley2003}
Hartley, R., Zisserman, A.: Multiple View Geometry in Computer Vision.
\newblock Cambridge University Press (2003)

\bibitem{Lowe2004}
Lowe, D.G.: Distinctive image features from scale-invariant keypoints.
\newblock Int. J. Comput. Vision \textbf{60}(2), 91--110 (2004)

\bibitem{Ma2004}
Ma, Y., Soatto, S., Kosecka, J., S., S.S.: An Invitation to {3-D} Vision: From
  Images to Geometric Models.
\newblock Springer (2004)

\bibitem{Mcnemar1947}
McNemar, Q.: Note on the sampling error of the difference between correlated
  proportions or percentages.
\newblock Psychometrika \textbf{12}(2), 153--157 (1947).
\newblock \doi{10.1007/BF02295996}

\bibitem{Neubeck2006}
Neubeck, A., Van~Gool, L.: Efficient non-maximum suppression.
\newblock In: Pattern Recognition, 2006. ICPR 2006. 18th International
  Conference on, vol.~3, pp. 850--855 (2006)

\bibitem{Nixon2002}
Nixon, M., Aguado, A.: Feature Extraction and Image Processing.
\newblock Electronics \& Electrical. Newnes (2002)

\bibitem{Ozkan2006}
Ozkan, D.: Feature selection for face recognition using a genetic algorithm.
\newblock Department of Computer Engineering, Bilkent University  (2006)

\bibitem{Perdoch2007}
Perdoch, M., Matas, J., Obdrzalek, S.: Stable affine frames on isophotes.
\newblock In: ICCV (2007)

\bibitem{Pospichal2010}
Pospichal, P., Jaros, J., Schwarz, J.: Parallel genetic algorithm on the cuda
  architecture.
\newblock In: Applications of Evolutionary Computation, \emph{Lecture Notes in
  Computer Science}, vol. 6024, pp. 442--451. Springer Berlin Heidelberg (2010)

\bibitem{Ripley1976}
Ripley, B.D.: The second order analysis of stationary point processes.
\newblock Journal of Applied Probability \textbf{13}, 255--266 (1976)

\bibitem{Ripley1977}
Ripley, B.D.: Modelling spatial patterns.
\newblock Journal of the Royal Statistics Society \textbf{39}, 172--212 (1977)

\bibitem{Rosten2010}
Rosten, E., Porter, R., Drummond, T.: Faster and better: {A} machine learning
  approach to corner detection.
\newblock IEEE Transactions on Pattern Analysis and Machine Intelligence
  \textbf{32}, 105--119 (2010)

\bibitem{Smith1997}
Smith, S., Brady, J.: Susan: A new approach to low level image processing.
\newblock International Journal of Computer Vision \textbf{23}(1), 45--78
  (1997)

\bibitem{Sola2007}
Sola, J.: Towards visual localization, mapping and moving objects tracking by a
  mobile robot: a geometric and probabilistic approach.
\newblock Ph.D. thesis, Institut National Politechnique de Toulouse (2007)

\bibitem{Still2005}
Still, M.: The Definitive Guide to {I}mage{M}agick (Definitive Guide).
\newblock Apress, Berkely, CA, USA (2005)

\bibitem{Student1908}
Student: The probable error of a mean.
\newblock Biometrika \textbf{6}(1), 1--25 (1908)

\bibitem{Tuytelaars2010}
Tuytelaars, T.: Dense interest points.
\newblock In: IEEE Conference on Computer Vision and Pattern Recognition
  (CVPR'10), pp. 2281--2288 (2010)

\end{thebibliography}

\end{document}